%% file: grasp-mm26-camera-ready.tex
\documentclass[sigconf]{acmart}

\usepackage{multirow}
\usepackage{colortbl}
\usepackage{pifont}
\usepackage{balance}

\input{figure.tex}

\AtBeginDocument{%
  }

\copyrightyear{2026}
\acmYear{2026}
\setcopyright{cc}
\setcctype{by}
\acmConference[MM '26]{Proceedings of the 34th ACM International Conference on Multimedia}{November 10--14, 2026}{Rio de Janeiro, Brazil}
\acmBooktitle{Proceedings of the 34th ACM International Conference on Multimedia (MM '26), November 10--14, 2026, Rio de Janeiro, Brazil}
\acmDOI{10.1145/3767308.3835872}
\acmISBN{979-8-4007-2213-4/2026/11}

\newcolumntype{C}[1]{>{\centering\arraybackslash}p{#1}}

\begin{document}

%%
%% The "title" command has an optional parameter,
%% allowing the author to define a "short title" to be used in page headers.
\title[GRASP for Fine-Grained Drone-View Cross-Modal Understanding]{GRASP: Granularity-Aware Region Alignment and Semantic Prototype Learning for Fine-Grained Cross-Modal Understanding in Drone Views}

%%
%% Each author is defined separately for ACM metadata extraction.
\author{Jiahui Cui}
\authornote{These authors contributed equally to this research.}
\email{cuijiahui24@mails.ucas.ac.cn}
\author{Yan Zhao}
\authornotemark[1]
\email{zhaoyan242@mails.ucas.ac.cn}
\author{Kan Wei}
\authornotemark[1]
\email{weikan24@mails.ucas.ac.cn}
\affiliation{%
  \institution{Aerospace Information Research Institute, Chinese Academy of Sciences}
  \city{Beijing}
  \country{China}}
\affiliation{%
  \institution{Key Laboratory of Target Cognition and Application Technology (TCAT)}
  \city{Beijing}
  \country{China}}
\affiliation{%
  \institution{University of Chinese Academy of Sciences}
  \city{Beijing}
  \country{China}}
\affiliation{%
  \institution{School of Electronic, Electrical and Communication Engineering, University of Chinese Academy of Sciences}
  \city{Beijing}
  \country{China}}

\author{Enze Zhu}
\email{zhuenze23@mails.ucas.ac.cn}
\affiliation{%
  \institution{Aerospace Information Research Institute, Chinese Academy of Sciences}
  \city{Beijing}
  \country{China}}
\affiliation{%
  \institution{Key Laboratory of Target Cognition and Application Technology (TCAT)}
  \city{Beijing}
  \country{China}}
\affiliation{%
  \institution{University of Chinese Academy of Sciences}
  \city{Beijing}
  \country{China}}
\affiliation{%
  \institution{School of Electronic, Electrical and Communication Engineering, University of Chinese Academy of Sciences}
  \city{Beijing}
  \country{China}}

\author{Peirong Zhang}
\email{zhangpeirong22@mails.ucas.ac.cn}
\affiliation{%
  \institution{Aerospace Information Research Institute, Chinese Academy of Sciences}
  \city{Beijing}
  \country{China}}
\affiliation{%
  \institution{Key Laboratory of Target Cognition and Application Technology (TCAT)}
  \city{Beijing}
  \country{China}}
\affiliation{%
  \institution{University of Chinese Academy of Sciences}
  \city{Beijing}
  \country{China}}
\affiliation{%
  \institution{School of Electronic, Electrical and Communication Engineering, University of Chinese Academy of Sciences}
  \city{Beijing}
  \country{China}}

\author{Lei Wang}
\email{wanglei002931@aircas.ac.cn}
\affiliation{%
  \institution{Aerospace Information Research Institute, Chinese Academy of Sciences}
  \city{Beijing}
  \country{China}}
\affiliation{%
  \institution{Key Laboratory of Target Cognition and Application Technology (TCAT)}
  \city{Beijing}
  \country{China}}

\author{Yiru Wang}
\correspondingauthor
\email{wangyiru@aircas.ac.cn}
\affiliation{%
  \institution{Aerospace Information Research Institute, Chinese Academy of Sciences}
  \city{Beijing}
  \country{China}}
\affiliation{%
  \institution{Key Laboratory of Target Cognition and Application Technology (TCAT)}
  \city{Beijing}
  \country{China}}

%%
%% The shortened author list prevents overlap in page headers.
\renewcommand{\shortauthors}{Jiahui Cui et al.}

%%
%% The abstract is a short summary of the work to be presented in the
%% article.
\begin{abstract}
Fine-grained cross-modal understanding in drone views is essential for aerial vision-language navigation. However, the inherent wide field of view and overhead perspective of drone scenarios impose dual challenges on vision-language understanding. At the macro level, overwhelming background clutter in visual representations leads to \textit{Cross-Modal Focus Misalignment}, where the model prioritizes global environmental similarities over specific object details. At the micro level, \textit{Visual Isomorphism} creates ambiguity, where candidates share similar geometric structures yet differ only in subtle attributes. To address these challenges, we propose the Granularity-Aware Region Alignment and Semantic Prototype (GRASP) learning framework, enhancing discriminative capability through two synergistic strategies. Specifically, we introduce Region-Focused Alignment (RFA) to promote object-centric cross-modal alignment while suppressing background interference. Concurrently, to tackle visual isomorphism, we propose Semantic Perturbation Enhanced Matching (SPEM), which leverages a foreground-purified Semantic Prototype Codebook (SPC) to construct semantically perturbed negatives for fine-grained semantic discrimination. Extensive experiments on the GeoText-1652 benchmark and the unseen ERA dataset demonstrate that GRASP achieves competitive performance in drone-view fine-grained image-text retrieval, validating its effectiveness for cross-modal understanding in aerial scenarios. Our code implementation is available at
\href{https://github.com/UCAS-JC/GRASP}{https://github.com/UCAS-JC/GRASP}.
\end{abstract}

%%
%% The code below is generated by the ACM CCS tool.


%%
%% Keywords. Separate the keywords with commas.

%%
%% This command processes the author and affiliation and title
%% information and builds the first part of the formatted document.
\maketitle

\motivation

\input{1_Introduction}
\input{2_Related_work}

\input{3_Methodology}
\input{4_Experiments}

\section{Conclusion}
In this paper, we propose GRASP framework for fine-grained cross-modal understanding in drone views. To tackle \textit{Cross-Modal Focus Misalignment}, we introduce RFA, enabling dual-granularity cross-modal alignment via region-level image-text contrastive learning and cross-modal ranking loss. To address \textit{Visual Isomorphism}, we propose SPEM, which leverages a foreground-purified SPC to construct hard negatives, forcing the model to discriminate fine-grained attribute differences. Employed exclusively during training, RFA and SPEM boost performance at no additional inference cost. Extensive experiments demonstrate that GRASP achieves competitive performance on the GeoText-1652 benchmark and remarkable zero-shot capability on the ERA dataset, validating its effectiveness for fine-grained cross-modal understanding in aerial scenarios.

%%
%% The acknowledgments section is defined using the "acks" environment.
\begin{acks}
This work was supported by the Key Laboratory of Target Recognition and Application under Grant E53712020F, the Key Program of the Chinese Academy of Sciences under Grant KGFZD-145-25-38 and RCJJ-145-24-13 and the Science and Disruptive Technology Program of Aerospace Information Research Institute under Grant AIRCAS2024-AIRCAS-SDTP-03.
\end{acks}

%%
%% The next two lines define the bibliography style to be used, and
%% the bibliography file.
\bibliographystyle{ACM-Reference-Format}
\balance
\bibliography{ijcai26}

\end{document}

%% file: figure.tex
\newcommand{\motivation}{
    \begin{figure}[!t] 
        \centering 
        \includegraphics[width=1.0\linewidth]{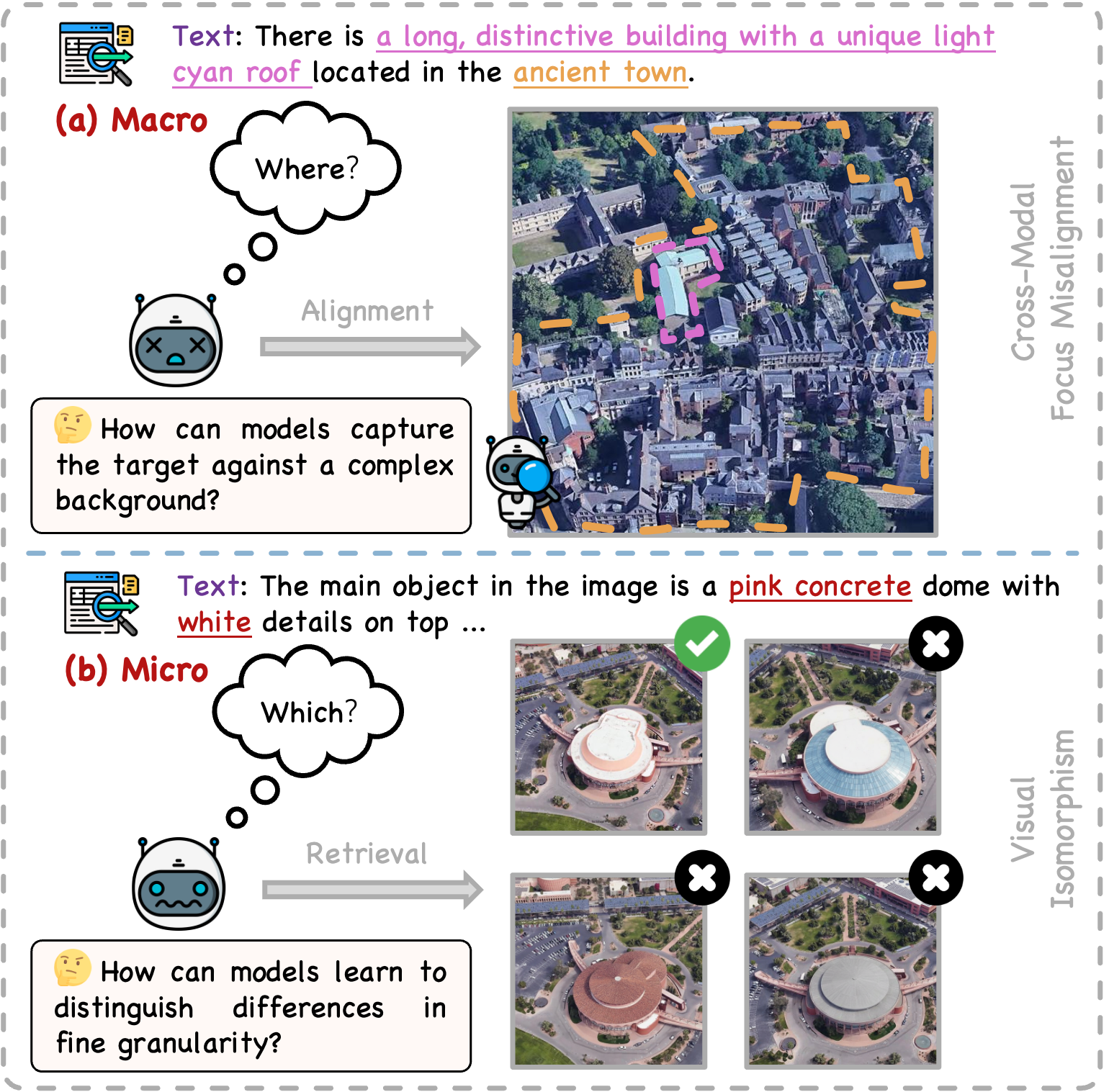}
        \caption{Dual challenges in drone-view understanding: (a) background-dominated alignment causes Cross-Modal Focus Misalignment; (b) geometrically similar objects with fine-grained attribute differences cause Visual Isomorphism.}
        \Description{Two panels illustrate cross-modal focus misalignment caused by background clutter and visual isomorphism among buildings with similar geometry but different textures.}
        \label{fig1}
    \end{figure}
}

\newcommand{\motivationtwo}{
    \begin{figure}[!t] 
        \centering 
        \includegraphics[width=1.0\linewidth]{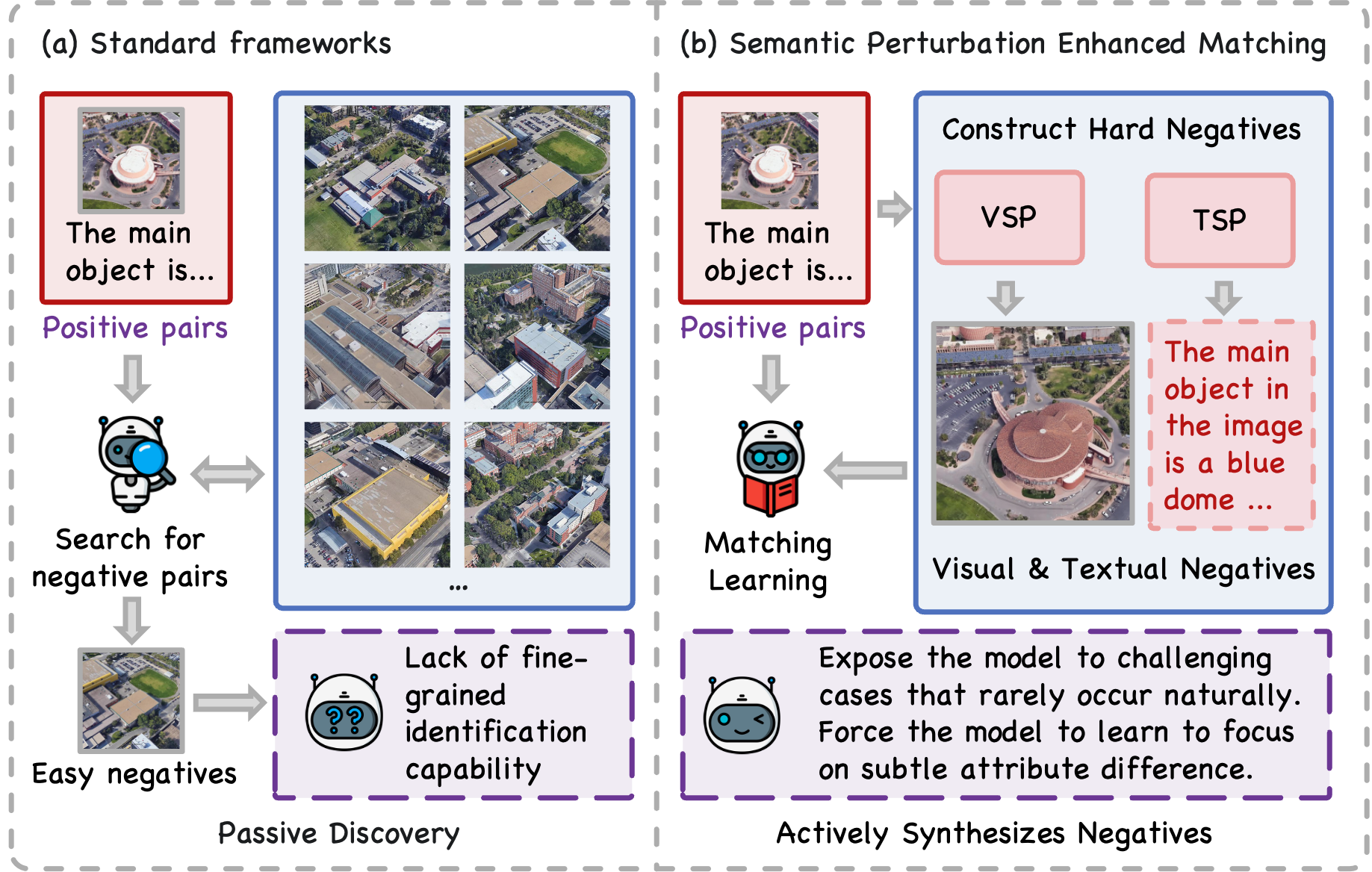}
        \caption{Standard passive in-batch sampling mainly yields easy negatives (a), whereas SPEM actively synthesizes hard textual and visual negatives via TSP and VSP (b) to improve attribute discrimination.}
        \Description{A comparison between passive in-batch negative sampling and SPEM, which synthesizes visual and textual hard negatives through VSP and TSP.}
        \label{fig2}
    \end{figure}
}

\newcommand{\model}{
    \begin{figure*}[!t]
    \centering
    \includegraphics[width=1.0\textwidth]{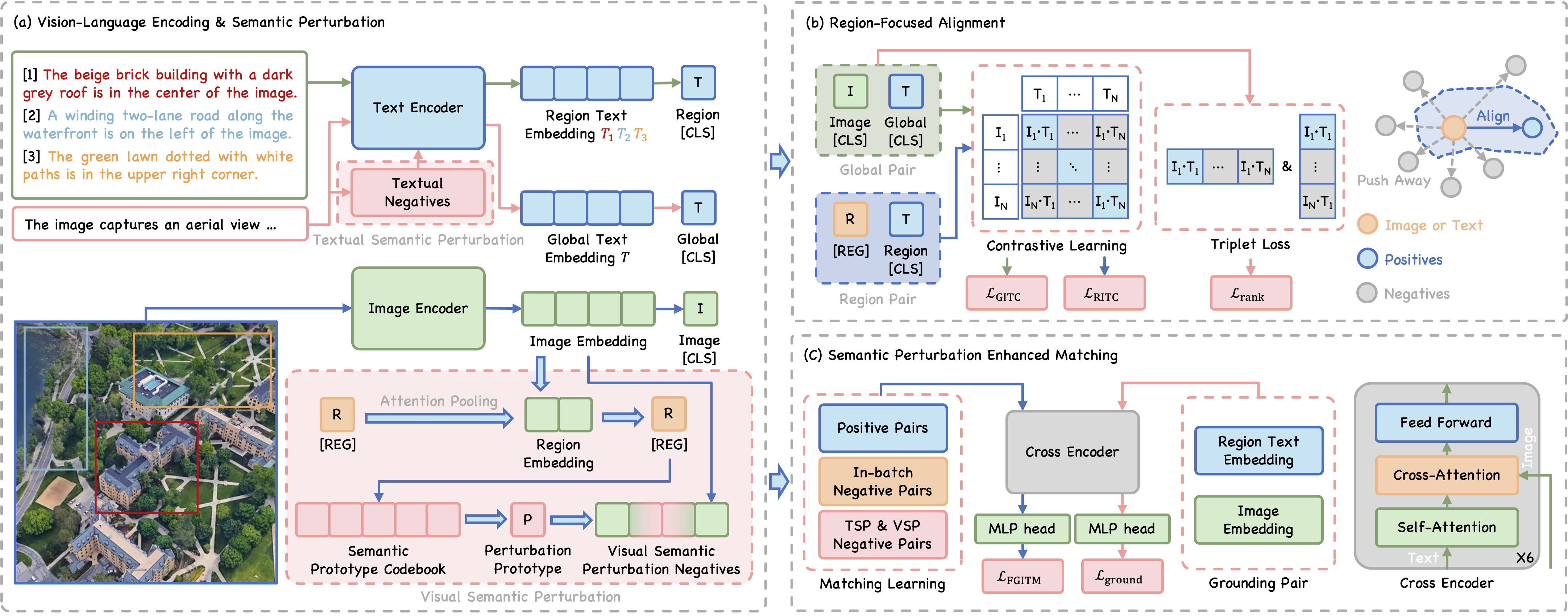}
    \caption{Overview of GRASP. (a) Encoders produce image and text embeddings, while TSP and SPC-based VSP synthesize hard negatives. (b) RFA combines global and region-level contrastive losses with ranking loss. (c) SPEM trains fine-grained matching with in-batch and synthesized negatives, together with grounding supervision.}
    \Description{The GRASP architecture combines vision-language encoding, textual and visual semantic perturbation, region-focused alignment, and fine-grained matching losses.}
    \label{fig3}
    \end{figure*}
}

\newcommand{\vsp}{
    \begin{figure}[!t] 
        \centering 
        \includegraphics[width=1.0\linewidth]{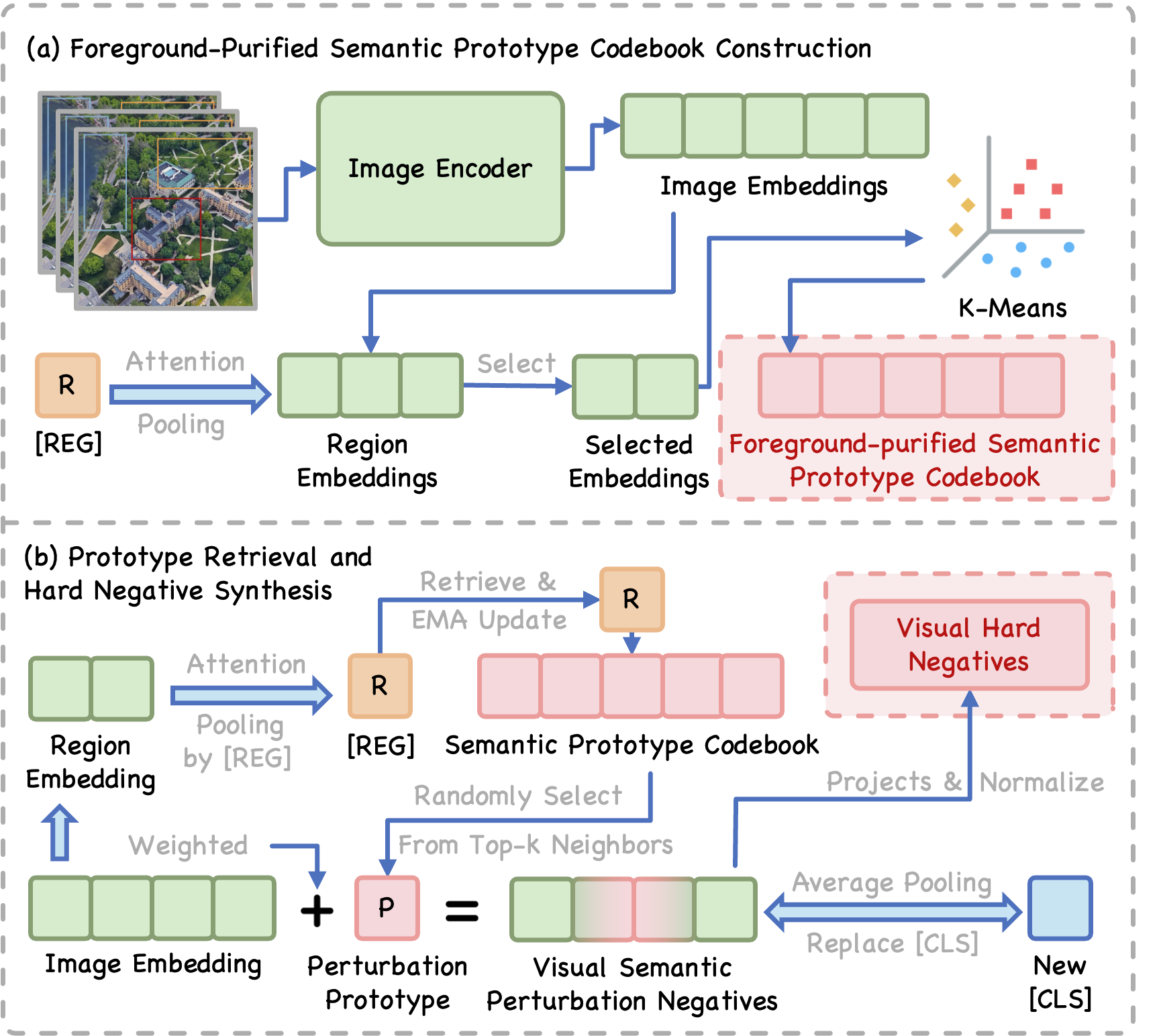}
        \caption{Visual Semantic Perturbation (VSP). (a) Foreground-Purified Semantic Prototype Codebook Construction: K-Means clusters attention-selected foreground patches into semantic prototypes. (b) Hard Negative Synthesis: a top-$k$ neighbor prototype is interpolated with the region embedding, while EMA updates the codebook.}
        \Description{VSP builds a semantic prototype codebook from attention-selected foreground patches and synthesizes visual hard negatives by retrieving and interpolating nearby prototypes.}
        \label{fig4}
    \end{figure}
}

\newcommand{\caseStudy}{
    \begin{figure}[!t] 
        \centering
        \includegraphics[width=0.88\linewidth]{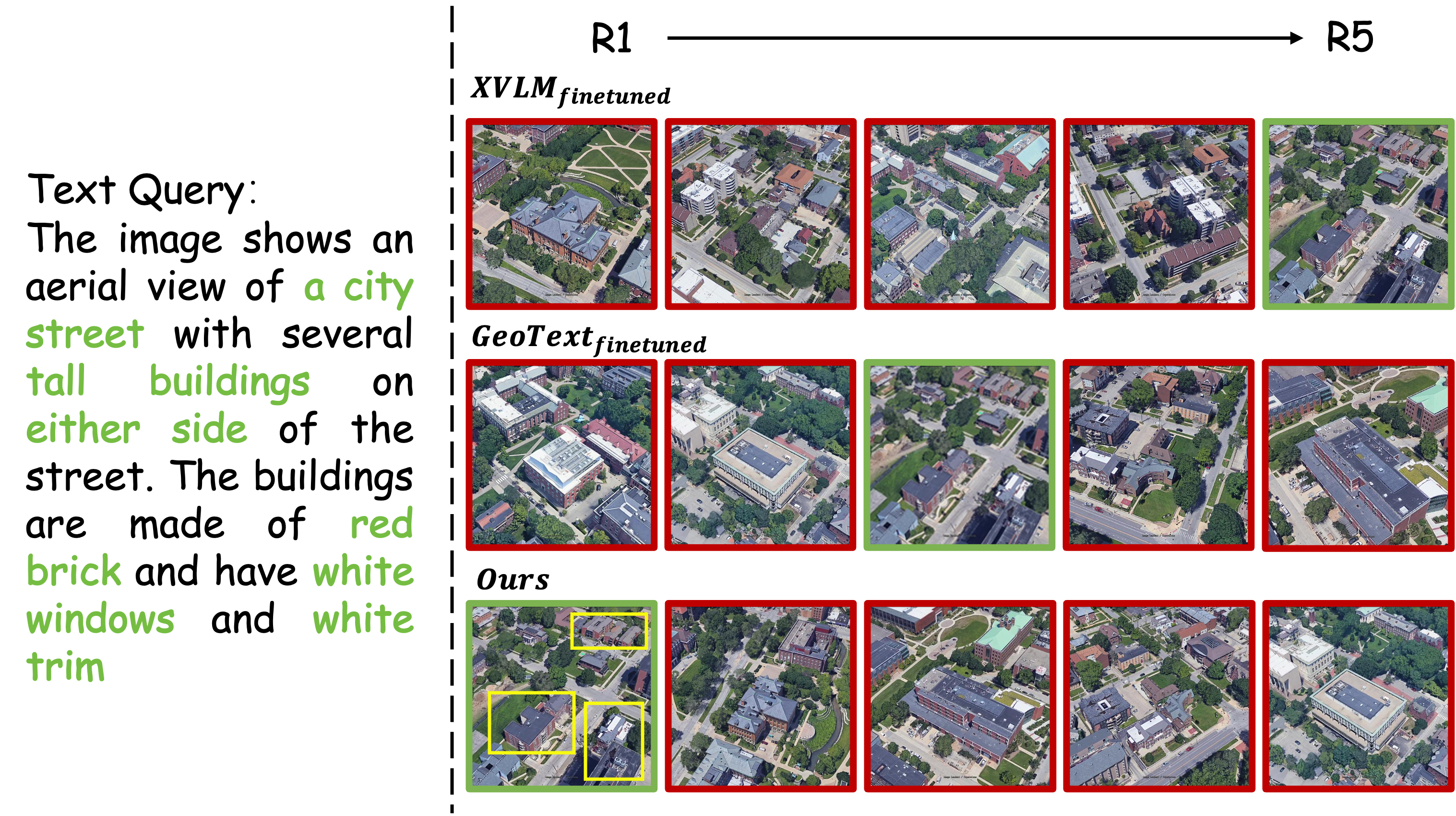}
        \caption{Qualitative text-to-image retrieval results ranked by similarity. Red and green boxes indicate incorrect and correct matches; query attributes are highlighted in green.
        }
        \Description{Ranked text-to-image retrieval results for two baselines and GRASP. Red boxes mark false matches, green boxes mark true matches, and query keywords are highlighted in green.}
        \label{caseStudy}
    \end{figure}
}

\newcommand{\Attention}{
    \begin{figure}[!t] 
        \centering 
        \includegraphics[width=0.9\linewidth]{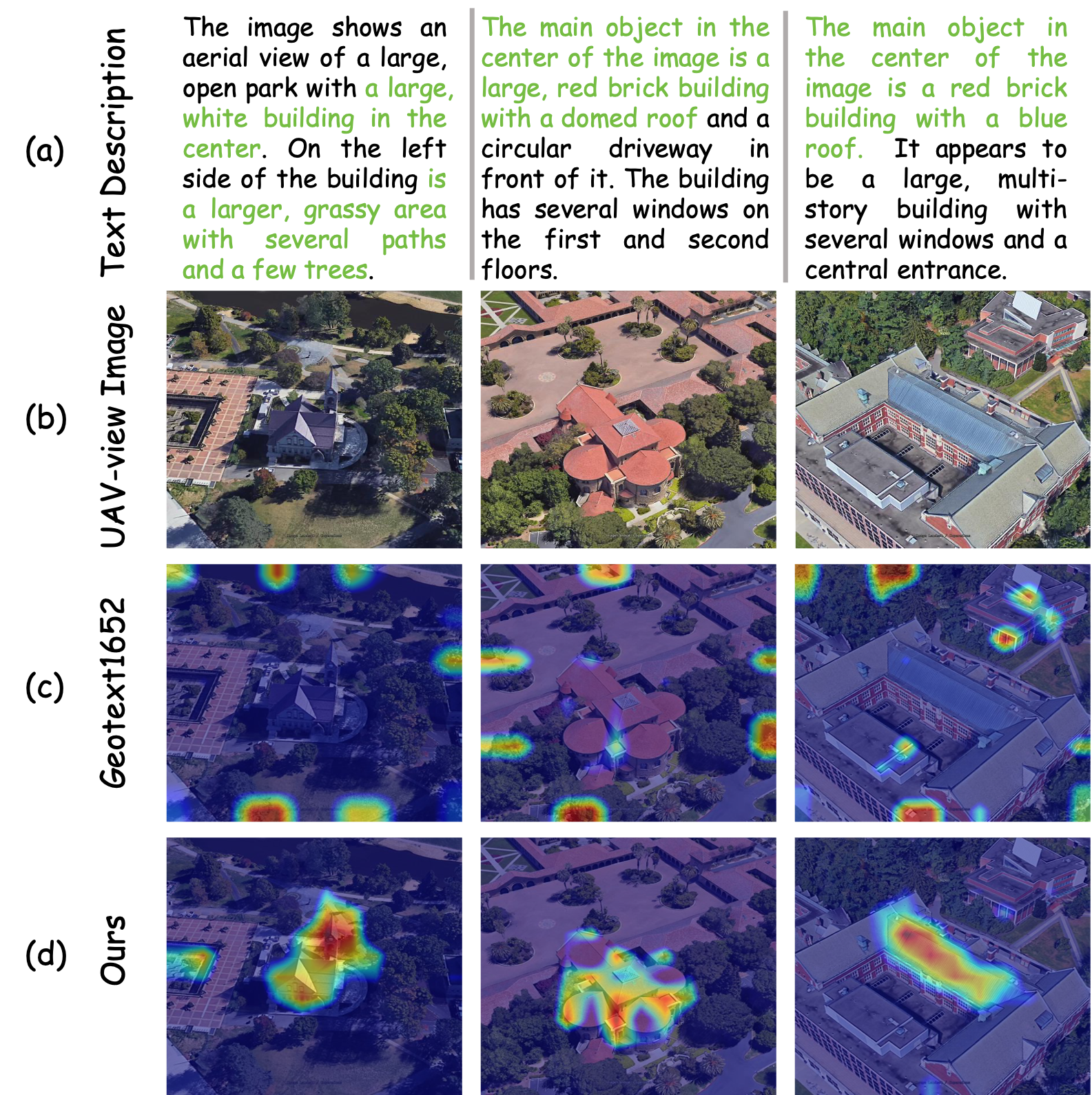}
        \caption{Visualization of activation maps. (a) Text descriptions with key descriptions highlighted in green. (b) UAV images. (c) GeoText-1652 baseline \cite{chu2024towards}. (d) GRASP.}
        \Description{Three examples compare text descriptions, UAV-view images, baseline activation maps, and GRASP activation maps; GRASP focuses more precisely on the described buildings.}
        \label{Attention}
    \end{figure}
}

%% file: 1_Introduction.tex
\section{Introduction}

Unmanned Aerial Vehicles (UAVs) have become essential platforms across diverse real-world applications. In the realm of sustainable agriculture, UAVs facilitate precision crop monitoring and targeted intervention \cite{UAVs_challenge,Towards_smart,Sustainable_agriculture}. In public safety contexts, they are indispensable for disaster response and search-and-rescue missions \cite{meguro2006disaster,rashid2020socialdrone,mehbodniya2022improving}. Similarly, within smart city frameworks, UAVs provide scalable solutions for traffic surveillance and facility inspection \cite{Real_time}. Propelled by recent strides in vision-language models, Natural Language-Guided Drones (NLGD) have emerged as a flexible paradigm for human-machine interaction \cite{chandarana2017fly,blukis2019learning,huang2019flight}. Central to this paradigm is the ability to interpret the semantic correspondence between natural language and aerial imagery, which necessitates deep cross-modal understanding within drone-view contexts.

Existing approaches use pre-trained vision-language models as feature extractors and align modalities through contrastive learning and image-text matching \cite{radford2021learning,li2021align,li2022blip}. GeoText-1652 \cite{chu2024towards} shifts drone-view retrieval from coarse alignment toward discriminative and object-centric reasoning. However, achieving fine-grained cross-modal understanding in drone views poses distinctive challenges at both macro and micro levels.

% Existing approaches typically employ pre-trained vision-language models for feature extraction and achieve cross-modal alignment through contrastive learning and image-text matching \cite{radford2021learning,li2021align,li2022blip}. Notably, the recent GeoText-1652 benchmark \cite{chu2024towards} necessitates a move towards fine-grained understanding, which shifts the paradigm from coarse alignment to discriminative and object-centric reasoning in drone-view retrieval. However, achieving fine-grained cross-modal understanding in drone views poses distinctive challenges at both macro and micro levels.

At the macro level, there exists a significant \textit{Cross-Modal Focus Misalignment}. Unlike ground-level photography where subjects are naturally framed, drone cameras capture expansive areas where targets are dwarfed by complex, noisy environments. Consequently, while textual descriptions provide elaborate attributes of specific targets, these objects visually occupy only a minute fraction of the wide field-of-view aerial imagery. As illustrated in Figure~\ref{fig1} (a), this semantic-visual disparity causes existing global image-text alignment methods to be inevitably overwhelmed by background noise, failing to ground the specific targets defined in the text \cite{chu2024towards}.

At the micro level, \textit{Visual Isomorphism} is uniquely pronounced in aerial scenes due to the top-down perspective. Because drones observe the world from above, critical vertical geometric cues—such as building facades, doors, and relative object heights that standard models rely heavily upon—are severely compressed or obscured. As a result, distinct objects often project deceptively similar structural layouts (Figure ~\ref{fig1} (b)). This heightened visual ambiguity forces the matching process to rely heavily on fine-grained discriminative attributes, including materials and colors. Unfortunately, standard contrastive and matching frameworks adopt passive in-batch negative sampling, which rarely encounters such visually isomorphic negative pairs in a natural batch, thus limiting the model's ability to learn these critical fine-grained differences \cite{zeng2021multi,chu2024towards}.

To address these challenges, we propose the Granularity-Aware Region Alignment and Semantic Prototype (GRASP) learning framework. GRASP enhances fine-grained cross-modal understanding through two synergistic strategies carefully tailored for drone views: Region-Focused Alignment (RFA) and Semantic Perturbation Enhanced Matching (SPEM). Specifically, to overcome the overwhelming background clutter, the RFA mechanism utilizes region-level image-text contrastive learning and a cross-modal ranking loss to explicitly enhance local, object-centric understanding while suppressing visual background interference. Concurrently, to tackle the severe visual homogenization caused by the overhead view, we propose SPEM. As illustrated in Figure~\ref{fig2}, unlike standard frameworks that rely on passive easy negatives, SPEM actively constructs semantically perturbed hard negatives via Textual Semantic Perturbation (TSP) and Visual Semantic Perturbation (VSP). In detail, TSP deterministically substitutes fine-grained attribute words to generate hard textual negatives while preserving the exact sentence context. Meanwhile, VSP establishes a foreground-purified Semantic Prototype Codebook (SPC) and leverages it to synthesize visual negatives that are strictly bounded within plausible attribute distributions, effectively averting out-of-distribution noise. TSP together with VSP explicitly forces the model to focus on the subtle attribute differences that distinguish visually isomorphic targets. Notably, RFA and SPEM are employed exclusively during the training phase. By bolstering the model's cross-modal understanding capabilities, they achieve significant performance gains with absolute zero additional computational overhead during inference.

Our main contributions are summarized as follows:
\begin{itemize}
    \item A Granularity-Aware Region Alignment and Semantic Prototype (GRASP) learning framework is presented, which systematically addresses the distinct cross-modal focus misalignment and visual isomorphism challenges in overhead drone-view understanding, introducing zero inference cost.
    \item A Region-Focused Alignment (RFA) mechanism is designed, which introduces region-level image-text contrastive learning and a cross-modal ranking loss to recalibrate alignment across both full-image and object-centric levels, mitigating background distraction.
    \item A Semantic Perturbation Enhanced Matching (SPEM) strategy is proposed, which explicitly constructs counterfactual hard negatives via controlled fine-grained attribute word replacement and visual prototype feature perturbation. This effectively forces the model to learn fine-grained semantic discrimination under severe visual isomorphism.
    \item Experiments on the GeoText-1652 benchmark and the unseen ERA dataset demonstrate that GRASP achieves competitive performance in drone-view fine-grained image-text retrieval, validating its robustness and generalizability for cross-modal understanding in aerial scenarios.
\end{itemize} 

%% file: 2_Related_work.tex
\section{Related Work}

\subsection{Vision-Language Cross-Modal Alignment}
Vision Language Models (VLMs) learn joint representations of images and text through large-scale pre-training. Early works like CLIP \cite{radford2021learning} employ contrastive learning for global image-text alignment. Subsequent dual-stream architectures improve upon this foundation: ALBEF \cite{li2021align} introduces hard negative mining with Image-Text Contrastive (ITC) loss, UNITER \cite{chen2020uniter} proposes word-region alignment, and XVLM \cite{zeng2021multi} incorporates multi-level alignment tasks. METER \cite{dou2022empirical} systematically evaluates encoder and fusion strategies, while BLIP \cite{li2022blip,li2023blip} unifies understanding and generation via lightweight modules. However, global alignment often fails to capture fine-grained semantic distinctions. To address this, hierarchical approaches encode semantic relationships using partial order constraints \cite{vendrov2015order,nguyen2017hierarchical} or hyperbolic spaces \cite{ganea2018hyperbolic,pal2024compositional}. Region-based methods align text with segmented image regions \cite{arbelaez2010contour,zhang2020self} or dynamically reweight hierarchical features \cite{wu2025mara}. Recent work \cite{wang2025enhancing} explores constructing negative samples to foster fine-grained alignment. However, this approach faces two limitations in aerial scenarios. First, it is easily overwhelmed by severe background noise, failing to produce targeted perturbations on the specific object. Second, its token-level manipulation disrupts region-level integrity, as it fragments targets that naturally span multiple tokens. Consequently, achieving fine-grained cross-modal understanding in drone views remains a significant challenge.

\subsection{Cross-Modal Localization and Navigation}
Cross-view localization \cite{zheng2020university,berton2022deep} addresses viewpoint-invariant visual matching, with transformer-based methods \cite{chen2023cross} attending to salient regions and partitioning strategies \cite{wang2021each,rodrigues2022global} enriching representations. However, these purely visual approaches lack language grounding capabilities. Vision Language Navigation (VLN) \cite{anderson2018vision} extends cross-modal understanding to embodied agents, requiring them to ground natural language instructions in visual observations for sequential decision-making. Indoor VLN methods achieve this through cross-modal attention \cite{majumdar2020improving,qi2021road}, episodic memory \cite{hong2021vln}, and structured spatial representations \cite{georgakis2022cross}. Recent efforts integrate natural language for aerial cross-modal understanding. CrossText2Loc \cite{ye2025cross} and UniLoc \cite{xia2024uniloc} enable text-based satellite retrieval. GeoText-1652 \cite{chu2024towards} establishes a drone-view benchmark with spatial matching, while VCSR \cite{huang2024visual} employs textual cues to guide semantic reasoning within the drone-view visual context.

% Unlike existing methods, our work focuses on addressing the issue of insufficient fine-grained alignment in drone views, which has been historically overlooked.

\motivationtwo
\model

%% file: 3_Methodology.tex
\section{Methodology}

\subsection{Vision-Language Encoding}
\label{sec:arch}

As illustrated in Figure~\ref{fig3}(a), we first perform vision-language encoding to obtain visual and textual embeddings that serve as the input representations for subsequent RFA and SPEM. Specifically, the Vision Encoder employs a Swin Transformer to encode an input image $I$ into a sequence of patch embeddings
\begin{equation*}
\mathbf{V} = \{\mathbf{v}_{\mathrm{cls}}, \mathbf{v}_1, \ldots, \mathbf{v}_N\} \in \mathbb{R}^{(N+1) \times d_{\mathrm{v}}},
\end{equation*}
where $N$ denotes the number of patches and $d_{\mathrm{v}} = 1024$. In parallel, the Text Encoder adopts BERT to encode an input text $T$ into a sequence of token embeddings $\mathbf{W} = \{\mathbf{w}_{\mathrm{cls}}, \mathbf{w}_1, \ldots, \mathbf{w}_L\} \in \mathbb{R}^{(L+1) \times d_{\mathrm{t}}}$, where $d_{\mathrm{t}} = 768$. 

For global contrastive learning in RFA, the global features $\mathbf{v}_{\mathrm{cls}}$ and $\mathbf{w}_{\mathrm{cls}}$ are projected to a joint embedding space $\mathbb{R}^{256}$ via linear projections, followed by L2 normalization for cosine similarity computation. To support fine-grained matching in SPEM, we partition the BERT encoder: the early layers are utilized for unimodal textual encoding, while the last six layers are dedicated to deep cross-modal interaction via cross-attention to capture complex visual-linguistic alignments. Structurally, RFA operates on the embeddings produced by the vision and text encoders, while SPEM further relies on the fused cross-modal representations.

\subsection{Region-Focused Alignment}
\label{sec:rfa}

The RFA module takes the visual and textual embeddings from Section~\ref{sec:arch} as input and establishes multi-granularity cross-modal correspondences by performing global/region image-text contrastive alignment and margin-based ranking against the hardest in-batch negatives, as depicted in Figure~\ref{fig3}(b).

\paragraph{Global Representation Learning.}
Given a batch of $B$ image-text pairs $\{(I_i, T_i)\}_{i=1}^{B}$, we first project the global features $\mathbf{v}_{\mathrm{cls}}$ and $\mathbf{w}_{\mathrm{cls}}$ into a shared embedding space and apply L2 normalization:
\begin{equation}
\hat{\mathbf{v}}_{\mathrm{cls}} = \frac{\phi_{\mathrm{v}}(\mathbf{v}_{\mathrm{cls}})}{\|\phi_{\mathrm{v}}(\mathbf{v}_{\mathrm{cls}})\|_2}, \quad \hat{\mathbf{w}}_{\mathrm{cls}} = \frac{\phi_{\mathrm{t}}(\mathbf{w}_{\mathrm{cls}})}{\|\phi_{\mathrm{t}}(\mathbf{w}_{\mathrm{cls}})\|_2},
\end{equation}
where $\phi_{\mathrm{v}}: \mathbb{R}^{d_{\mathrm{v}}} \rightarrow \mathbb{R}^{256}$ and $\phi_{\mathrm{t}}: \mathbb{R}^{d_{\mathrm{t}}} \rightarrow \mathbb{R}^{256}$ are learnable linear projections. The global similarity matrix $\mathbf{S} \in \mathbb{R}^{B \times B}$ is computed as:
\begin{equation}
s_{ij} = \cos(\hat{\mathbf{v}}_{\mathrm{cls}}^{(i)}, \hat{\mathbf{w}}_{\mathrm{cls}}^{(j)}) = \hat{\mathbf{v}}_{\mathrm{cls}}^{(i)\top} \hat{\mathbf{w}}_{\mathrm{cls}}^{(j)}.
\end{equation}
We optimize global alignment via a symmetric InfoNCE objective:
\begin{equation}
\begin{split}
\mathcal{L}_{\mathrm{GITC}} = -\frac{1}{2B}\sum_{i=1}^{B}\bigg[&\log\frac{\exp(s_{ii}/\tau)}{\sum_{j=1}^{B}\exp(s_{ij}/\tau)} \\
&+ \log\frac{\exp(s_{ii}/\tau)}{\sum_{j=1}^{B}\exp(s_{ji}/\tau)}\bigg]
\end{split}
\end{equation}
where $\tau$ denotes the temperature.

\paragraph{Region-Level Representation Learning.}
To capture fine-grained target semantics, we introduce a learnable region query vector $\mathbf{q}_{\mathrm{reg}} \in \mathbb{R}^{d_{\mathrm{v}}}$ that dynamically aggregates visual patch features within the target bounding box. Let $\mathbf{V}_{\mathrm{bbox}} = \{\mathbf{v}_1, \mathbf{v}_2, \ldots, \mathbf{v}_M\} \in \mathbb{R}^{M \times d_{\mathrm{v}}}$ denote the subset of $M$ patch features spatially located within the bounding box. The region representation is computed via scaled dot-product attention pooling:
\begin{equation}
\mathbf{a} = \operatorname{Softmax}\left(\frac{\mathbf{q}_{\mathrm{reg}}\mathbf{V}_{\mathrm{bbox}}^\top}{\sqrt{d_{\mathrm{v}}}}\right), \quad \mathbf{r} = \mathbf{a}\mathbf{V}_{\mathrm{bbox}},
\end{equation}
where $\mathbf{a} \in \mathbb{R}^{M}$ represents the attention weights over patches. The aggregated region feature $\mathbf{r}$ is projected and normalized as $\hat{\mathbf{r}} = \phi_{\mathrm{r}}(\mathbf{r}) / \|\phi_{\mathrm{r}}(\mathbf{r})\|_2$ with $\phi_{\mathrm{r}}: \mathbb{R}^{d_{\mathrm{v}}} \rightarrow \mathbb{R}^{256}$. We then compute the region-text similarity $s^{\mathrm{r}}_{ij} = \cos(\hat{\mathbf{r}}^{(i)}, \hat{\mathbf{w}}_{\mathrm{cls}}^{(j)})$ and optimize:
\begin{equation}
\begin{split}
\mathcal{L}_{\mathrm{RITC}} = -\frac{1}{2B}\sum_{i=1}^{B}\bigg[&\log\frac{\exp(s^{\mathrm{r}}_{ii}/\tau)}{\sum_{j=1}^{B}\exp(s^{\mathrm{r}}_{ij}/\tau)} \\
&+ \log\frac{\exp(s^{\mathrm{r}}_{ii}/\tau)}{\sum_{j=1}^{B}\exp(s^{\mathrm{r}}_{ji}/\tau)}\bigg]
\end{split}
\end{equation}

\paragraph{Hard Negative Mining with Margin Constraint.}
To further enhance discriminability, we enforce a margin-based ranking constraint that explicitly separates positive pairs from the hardest in-batch negatives. For each sample, we define such negatives as:
\begin{equation}
s_{\mathrm{neg}}^{\mathrm{i2t}} = \max_{j \neq i} s_{ij}, \quad s_{\mathrm{neg}}^{\mathrm{t2i}} = \max_{j \neq i} s_{ji}
\end{equation}
The ranking objective ensures that the positive similarity $s_{\mathrm{pos}} = s_{ii}$ exceeds these hard negatives by at least margin $m$:
\begin{equation}
\begin{split}
\mathcal{L}_{\mathrm{rank}} = \frac{1}{2}\big[&\max(0, m - s_{\mathrm{pos}} + s_{\mathrm{neg}}^{\mathrm{i2t}}) \\
&+ \max(0, m - s_{\mathrm{pos}} + s_{\mathrm{neg}}^{\mathrm{t2i}})\big]
\end{split}
\end{equation}

\subsection{Semantic Perturbation Enhanced Matching}

\vsp

\label{sec:spem}

SPEM actively synthesizes semantically perturbed hard negatives in both textual and visual modalities, as illustrated in Figure~\ref{fig2} and Figure~\ref{fig3} (a). Unlike passive in-batch mining that is bottlenecked by dataset distribution, SPEM constructs counterfactual samples that preserve overall structure while perturbing discriminative attributes, exposing the model to isomorphic challenges.

\paragraph{Textual Semantic Perturbation.}
TSP generates hard textual negatives by systematically replacing attribute words in the input text $T$. Instead of relying on generative language models for text perturbation, which might introduce unpredictable syntactic changes and hallucinations, we adopt a deterministic, hierarchical dictionary-based substitution. From a causal representation learning perspective, this guarantees \textit{strict control variates}: it modifies \textit{only} the targeted discriminative attribute while preserving the exact grammatical structure and contextual semantics. This precision is crucial for constructing high-quality counterfactuals. Given a text with $L$ tokens $\{w_1, w_2, \ldots, w_L\}$, we construct a perturbed text $\tilde{T}$ through three priority levels:

\subparagraph{Level 1 (Primary Attributes):} We first identify color words $\mathcal{W}_{\mathrm{c}}$ and orientation words $\mathcal{W}_{\mathrm{o}}$ that serve as primary discriminative attributes for building objects in aerial views. For each $w_i \in \mathcal{W}_{\mathrm{c}} \cup \mathcal{W}_{\mathrm{o}}$, we retrieve its semantic opposite from predefined dictionaries:
\begin{equation}
\tilde{w}_i = \mathcal{D}_{\mathrm{opposite}}(w_i), \quad \forall w_i \in \mathcal{W}_{\mathrm{c}} \cup \mathcal{W}_{\mathrm{o}}
\end{equation}
This targets the most salient attributes that distinguish visually similar buildings.

\subparagraph{Level 2 (Antonym Substitution):} When the number of replaced words $|\tilde{\mathcal{W}}| < \eta$ falls below threshold $\eta$, we expand to general antonyms. For remaining tokens, we query an antonym dictionary:
\begin{equation}
\tilde{w}_j = \mathcal{D}_{\mathrm{antonym}}(w_j), \quad \text{if } w_j \in \operatorname{dom}(\mathcal{D}_{\mathrm{antonym}})
\end{equation}

\subparagraph{Level 3 (Random Perturbation):} If still insufficient, random words with length greater than three are selected for replacement to reach the preset threshold $\eta$:
\begin{equation}
\tilde{w}_k = \mathcal{R}(\mathcal{V}), \quad \text{where } |w_k| > 3
\end{equation}
where $\mathcal{R}(\mathcal{V})$ samples from the vocabulary. This hierarchical design ensures that perturbations prioritize semantically meaningful attributes while maintaining sufficient diversity.

\paragraph{Visual Semantic Perturbation.}
VSP constructs hard visual negatives in feature space through attribute-level perturbation, as illustrated in Figure~\ref{fig4}. Unlike prior methods that construct visual dictionaries from generic scenes, which introduce severe background noise irrelevant to the structural perturbations of the target objects, we propose a Semantic Prototype Codebook (SPC) that captures discrete, foreground-purified semantic prototypes.

\subparagraph{Foreground-Purified Codebook Construction.}
As depicted in Figure~\ref{fig4} (a), the SPC $\mathcal{C} = \{\mathbf{c}_1, \mathbf{c}_2, \ldots, \mathbf{c}_K\}$ is a discrete set of $K$ semantic attribute prototypes, where $\mathbf{c}_k \in \mathbb{R}^{d_{\mathrm{v}}}$. To construct foreground-purified prototypes, we leverage the attention mechanism from RFA. For each training image, the attention weights over bounding box patches are computed as:
\begin{equation}
\mathbf{a} = \operatorname{Softmax}\left(\frac{\mathbf{q}_{\mathrm{reg}}\mathbf{V}_{\mathrm{bbox}}^\top}{\sqrt{d_{\mathrm{v}}}}\right)
\end{equation}
We retain only patches with attention weights in the top $\rho\,\%$ to form the foreground feature set, filtering out background clutter:
\begin{equation}
\mathcal{F} = \{\mathbf{v}_i \mid a_i > \operatorname{Percentile}(\mathbf{a}, 100 - \rho)\}
\end{equation}
The foreground features from all training samples $\bigcup \mathcal{F}$ are clustered via K-Means, producing $K$ cluster centroids as semantic prototypes that represent pure textural attributes of target objects.

\subparagraph{Prototype Retrieval and Perturbation.}
As shown in Figure~\ref{fig4} (b), for a region feature $\mathbf{r}$, we first retrieve its nearest neighbor prototype from SPC via cosine similarity:
\begin{equation}
\mathbf{c}_{\mathrm{nn}} = \operatorname*{arg\,max}_{\mathbf{c} \in \mathcal{C}} \frac{\mathbf{r}^\top \mathbf{c}}{\|\mathbf{r}\| \|\mathbf{c}\|}
\end{equation}
To generate semantically similar yet distinct perturbations, we sample a perturbation prototype $\mathbf{c}_{\mathrm{pert}}$ from the top-$k$ nearest neighbors of $\mathbf{c}_{\mathrm{nn}}$, excluding $\mathbf{c}_{\mathrm{nn}}$ itself:
\begin{equation}
\mathbf{c}_{\mathrm{pert}} \sim \operatorname{Uniform}(\mathcal{N}_k(\mathbf{c}_{\mathrm{nn}}) \setminus \{\mathbf{c}_{\mathrm{nn}}\})
\end{equation}
where $\mathcal{N}_k(\mathbf{c}_{\mathrm{nn}}) = \{\mathbf{c} \in \mathcal{C} : \operatorname{rank}(\cos(\mathbf{c}_{\mathrm{nn}}, \mathbf{c})) \leq k\}$. This sampling strategy ensures the perturbation remains within a plausible semantic neighborhood.

\subparagraph{Feature-Space Hard Negative Synthesis.}
For the foreground patch set $\mathcal{F}$ identified by attention weights, we generate perturbed features via weighted interpolation:
\begin{equation}
\mathbf{v}'_i = (1 - \alpha)\mathbf{v}_i + \alpha \mathbf{c}_{\mathrm{pert}}, \quad \forall \mathbf{v}_i \in \mathcal{F}
\end{equation}
where $\alpha \in [0, 1]$ is the mixing coefficient controlling perturbation strength. Background patches remain fixed to preserve scene context, ensuring that perturbations are exclusively applied to the target attributes. The perturbed patch sequence is used to recompute the global representation:
\begin{equation}
\mathbf{v}'_{\mathrm{cls}} = \operatorname{AvgPool}(\{\mathbf{v}'_1, \mathbf{v}'_2, \ldots, \mathbf{v}'_N\})
\end{equation}
which is projected to the joint space $\mathbb{R}^{256}$ and serves as a visual hard negative.

\subparagraph{Manifold-Constrained Perturbation.}
While feature-level perturbation often raises concerns about generating out-of-distribution (OOD) features, VSP intrinsically avoids this risk. Given that SPC is constructed exclusively from in-distribution foreground features (Eq. 12), and $\mathbf{c}_{\mathrm{pert}}$ is strictly constrained within the top-$k$ semantic neighborhood (Eq. 14), the convex combination (Eq. 15) guarantees that the synthesized negative features remain safely bounded within the local in-distribution manifold.

\subparagraph{Dynamic Codebook Update.}
During training, SPC is dynamically updated to track the evolving feature space. For each region feature $\mathbf{r}$, its nearest prototype is updated via Exponential Moving Average (EMA):
\begin{equation}
\mathbf{c}_{\mathrm{nn}} \leftarrow \lambda \mathbf{c}_{\mathrm{nn}} + (1 - \lambda) \operatorname{sg}(\mathbf{r})
\end{equation}
where $\lambda$ is the EMA momentum and $\operatorname{sg}(\cdot)$ stops gradients.

\paragraph{Cross-Modal Matching with Synthesized Negatives.}
As shown in Figure~\ref{fig3}(c), the cross-modal fusion encoder performs fine-grained matching by receiving five types of image-text pairs:
\begin{enumerate}
\item positive pairs $(I, T)$,
\item in-batch hard image negatives $(I^-_{\mathrm{hard}}, T)$ where
\begin{displaymath}
I^-_{\mathrm{hard}} = \operatorname*{arg\,max}_{j \neq i} s_{ji},
\end{displaymath}
\item in-batch hard text negatives $(I, T^-_{\mathrm{hard}})$ where
\begin{displaymath}
T^-_{\mathrm{hard}} = \operatorname*{arg\,max}_{j \neq i} s_{ij},
\end{displaymath}
\item TSP-generated text negatives $(I, \tilde{T})$, and
\item VSP-generated visual negatives $(\tilde{I}, T)$.
\end{enumerate}
The matching score $p(y\mid I, T)$ is computed through deep cross-attention, and the model is optimized via binary cross-entropy:
\begin{equation}
\mathcal{L}_{\mathrm{FGITM}} = -\mathbb{E}\big[y\log p + (1-y)\log(1-p)\big]
\end{equation}
where $y=1$ for positive pairs $(I, T)$, and $y=0$ for negative pairs $(I', T') \in \{(I^-_{\mathrm{hard}}, T), (I, T^-_{\mathrm{hard}}), (I, \tilde{T}), (\tilde{I}, T)\}$. Unlike standard InfoNCE contrastive losses that operate on globally pooled vectors and are inherently insensitive to micro-level perturbations (e.g., altering a single color word), we specifically employ BCE for $\mathcal{L}_{\text{FGITM}}$. Supported by the fusion encoder's deep cross-attention, BCE effectively supervises dense, token-to-patch interactions, explicitly penalizing the model for localized semantic mismatches.

\subsection{Training Objective}
\label{sec:objective}

\paragraph{Grounding Loss.}
For consistency with standard practice, we retain the grounding loss to supervise region-text alignment. Given a text segment $T_k$ and its corresponding ground-truth bounding box $b_k$, a bounding box regression head $H_{box}$ predicts $\hat{b}_k$ from the fused features. The grounding loss penalizes the discrepancy between predicted and ground-truth boxes through a combination of L1 distance and Generalized IoU loss:
\begin{equation}
\mathcal{L}_{\mathrm{ground}} = \sum_{k}\big[\lambda_{\mathrm{L1}}\|b_k - \hat{b}_k\|_1 + \lambda_{\mathrm{giou}}\mathcal{L}_{\mathrm{GIoU}}(b_k, \hat{b}_k)\big]
\end{equation}
where $\lambda_{\mathrm{L1}}$ and $\lambda_{\mathrm{giou}}$ are the weights for L1 and GIoU losses.

\paragraph{Overall Objective.}
The complete training objective is a weighted sum of all component losses:
\begin{equation}
\begin{split}
\mathcal{L} = &\lambda_1\mathcal{L}_{\mathrm{GITC}} + \lambda_2\mathcal{L}_{\mathrm{RITC}} + \lambda_3\mathcal{L}_{\mathrm{rank}} \\
&+ \lambda_4\mathcal{L}_{\mathrm{FGITM}} + \lambda_5\mathcal{L}_{\mathrm{ground}}
\end{split}
\end{equation}
where $\lambda_1, \lambda_2, \lambda_3, \lambda_4, \lambda_5$ are the weight coefficients for each loss term.

%% file: 4_Experiments.tex
\section{Experiments}

\begin{table}[!t] % 跨双栏表格
\centering
\caption{Zero-shot and fine-tuned retrieval on GeoText-1652 benchmark. Results are presented using Recall@K (R@K) and mean Recall (mR) for both Image Query and Text Query tasks. The best performances are in \textbf{bold}.}
\label{tab:main_results}

% 使用 tabular* 并设置宽度为 \textwidth，配合 @{\extracolsep{\fill}} 实现列宽自动调整
% \tiny
\resizebox{\columnwidth}{!}{%
\begin{tabular}{l c C{1.15cm} c c c c c c c}
\toprule
\multirow{2}{*}{\textbf{Method}} & \multirow{2}{*}{\textbf{Params}} & 
\multirow{2}{*}{\shortstack[c]{\textbf{Pretrained}\\\textbf{Images}}}
% \multirow{2}{*}{\textbf{Pretrained Images}}

% & 
% \multirow{2}{*}{%
% \parbox[c]{0.55cm}{\centering\textbf{Pretrained Images}}%
% } 

& \multicolumn{3}{c}{\textbf{Image Query (\%)}} & \multicolumn{3}{c}{\textbf{Text Query (\%)}} & \multirow{2}{*}{\textbf{mR (\%)}} \\
\cmidrule(lr){4-6} \cmidrule(lr){7-9}
 & & & R@1 & R@5 & R@10 & R@1 & R@5 & R@10 & \\
\midrule
% 第一部分子标题
\rowcolor{gray!20}
\multicolumn{10}{c}{\textit{Zero-Shot Evaluation on GeoText-1652}} \\
\midrule
UNITER\cite{chen2020uniter} & 300M & 4M & 2.5 & 7.4 & 11.8 & 0.9 & 2.7 & 4.2 & 4.92 \\
METER-Swin\cite{dou2022empirical} & 380M & 4M & 2.7 & 8.0 & 12.2 & 1.3 & 3.9 & 5.8 & 5.65 \\
ALBEF\cite{li2021align} & 210M & 4M & 2.9 & 8.1 & 12.4 & 1.8 & 4.8 & 7.1 & 6.18 \\
ALBEF\cite{li2021align}  & 210M & 14M & 3.0 & 9.1 & 14.2 & 1.1 & 3.5 & 5.3 & 6.03 \\
XVLM\cite{zeng2021multi} & 216M & 4M & 4.9 & 14.2 & 21.1 & 4.3 & 9.1 & 13.2 & 11.13 \\
XVLM\cite{zeng2021multi} & 216M & 16M & 5.0 & 14.4 & 21.4 & 4.5 & 9.9 & 13.4 & 11.43 \\
\addlinespace
% 第二部分子标题
\rowcolor{gray!20}
\multicolumn{10}{c}{\textit{Fine-Tuned Evaluation on GeoText-1652}} \\
\midrule
UNITER\cite{chen2020uniter} & 300M & 4M & 21.4 & 43.4 & 59.5 & 10.6 & 20.4 & 26.1 & 30.23 \\
METER-Swin\cite{dou2022empirical}  & 380M & 4M & 22.7 & 46.3 & 60.7 & 11.3 & 21.5 & 27.3 & 31.63 \\
ALBEF\cite{li2021align}  & 210M & 4M & 22.9 & 49.5 & 62.3 & 12.3 & 22.8 & 28.6 & 33.07 \\
ALBEF\cite{li2021align}  & 210M & 14M & 23.2 & 49.7 & 62.4 & 12.5 & 22.8 & 28.5 & 33.18 \\
XVLM\cite{zeng2021multi}  & 216M & 4M & 23.6 & 50.0 & 63.2 & 13.1 & 23.5 & 29.2 & 33.77 \\
XVLM\cite{zeng2021multi}  & 216M & 16M & 25.0 & 52.3 & 65.1 & 13.2 & 23.7 & 29.6 & 34.82 \\
GeoText-1652\cite{chu2024towards}  & 217M & 16M & 26.3 & 53.7 & 66.9 & 13.6 & 24.6 & 31.2 & 36.05 \\
NAS\cite{wang2025enhancing}  & 217M & 16M & 26.8 & 54.0 & 66.8 & 13.4 & 24.6 & 31.3 & 36.15 \\
\textbf{GRASP(ours)} & 217M & 16M & \textbf{29.6} & \textbf{57.0} & \textbf{69.7} & \textbf{14.5} & \textbf{25.4} & \textbf{31.8} & \textbf{38.04} \\
\bottomrule
\end{tabular}%
}
\end{table}

\subsection{Dataset and Evaluation Metrics}
\label{sec:dataset}
\paragraph{Dataset.}
We evaluate GRASP on two benchmarks: GeoText-1652 \cite{chu2024towards} for fine-tuned evaluation and ERA \cite{huang2024visual} for zero-shot transfer evaluation. GeoText-1652 is designed for fine-grained drone-view geo-localization with natural language. Following the official protocol, we use 701 locations for training, 351 for validation, and 600 for testing. ERA \cite{huang2024visual} is an aerial image-text retrieval benchmark containing 2,864 drone-view images. We use this dataset for zero-shot evaluation to assess cross-domain generalization capability.

\paragraph{Evaluation Metrics.}
% We adopt Recall@K (R@1, R@5, R@10) and mean Recall (mR) as evaluation metrics. Recall@K measures the percentage of queries for which the correct target appears within the top-K retrieved results. Mean Recall (mR) averages all Recall@K scores to provide an overall performance indicator. We evaluate both Image Query (image-to-text retrieval) and Text Query (text-to-image retrieval) tasks.
We report Recall@K (R@1, R@5, and R@10) and mean Recall (mR) for image-to-text and text-to-image retrieval; mR averages the six Recall@K scores.

\subsection{Implementation Details}
\label{sec:impl}

For fair comparison, we follow the GeoText-1652 setup \cite{chu2024towards}, which uses a standard XVLM \cite{zeng2021multi} model pre-trained on 16M image-text pairs as the backbone. We fine-tune our model on GeoText-1652 dataset. The SPC contains $K=2048$ prototypes. For SPC initialization, we retain the top $\rho=95\,\%$ patches by attention weights. The EMA momentum coefficient $\lambda$ is set to 0.995. For VSP, the mixing coefficient $\alpha$ is 0.7 and we sample perturbation prototypes from top-3 neighbors. For TSP, we replace up to 7 attribute words per sentence. The loss weights are set to $\lambda_1=\lambda_2=1.0$, $\lambda_3=\lambda_4=0.6$ and $\lambda_5=0.1$. The margin $m$ in ranking loss is 0.25. We use AdamW optimizer with learning rate $3\times10^{-5}$ and batch size 32.

\paragraph{Two-Stage Training Policy \& Zero Inference Cost.}
Fine-tuning adopts a two-stage procedure. The first stage performs warm-up training for 1 epoch, utilizing all five loss functions with partial FGITM that includes only positive pairs and in-batch negatives, without TSP/VSP-generated hard negatives. After this stage, foreground features are extracted using the trained model to initialize SPC. The second stage introduces SPEM and trains for 2 epochs, incorporating TSP and VSP-generated hard negatives into FGITM for fine-grained discrimination refinement. Crucially, SPC construction and hard-negative synthesis are used only during training and add no inference cost.

\begin{table}[!t]
\centering
\caption{Ablation Study on the GeoText-1652 benchmark.}
\label{tab:ablation_with_mr}
% 字体极小
\tiny
% 紧凑列间距
\setlength{\tabcolsep}{2pt}
\resizebox{\columnwidth}{!}{%
\begin{tabular}{cccccccccccc} % 12列
\toprule
% --- 表头第一行 ---
\multicolumn{4}{c}{\textbf{Components}} & \multicolumn{3}{c}{\textbf{Image Query(\%)}} & \multicolumn{3}{c}{\textbf{Text Query(\%)}} & \multirow{2}{*}{\textbf{mR(\%)}} \\
\cmidrule(lr){1-4} \cmidrule(lr){5-7} \cmidrule(lr){8-10}
% --- 表头第二行 ---
$\mathcal{L}_{\mathrm{rank}}$ & $\mathcal{L}_{\mathrm{RITC}}$ & TSP & VSP & R@1 & R@5 & R@10 & R@1 & R@5 & R@10 & \\
\midrule
& & & & 26.3 & 53.7 & 66.9 & 13.6 & 24.6 & 31.2 & 36.05 \\
 & & \checkmark & \checkmark & 28.4 & 55.7 & 68.2 & 13.9 & 25.0 & 31.2 & 37.01 \\
\checkmark & & \checkmark & \checkmark & 28.7 & 55.9 & 68.8 & 14.0 & 25.2 & 31.2 & 37.30 \\
 & \checkmark & \checkmark & \checkmark & 28.5 & 55.8 & 68.7 & 14.1 & 25.2 & 31.3 & 37.26 \\
\checkmark & \checkmark & & & 27.8 & 54.8 & 67.9 & 14.0 & 24.8 & 31.2 & 36.75 \\
\checkmark & \checkmark & \checkmark & & 28.4 & 56.0 & 68.9 & 14.0 & 25.2 & 31.5 & 37.33 \\
\checkmark & \checkmark & & \checkmark & 28.6 & 56.1 & 69.1 & 14.2 & 25.4 & 31.8 & 37.53 \\
\checkmark & \checkmark & \checkmark & \checkmark & \textbf{29.6} & \textbf{57.0} & \textbf{69.7} & \textbf{14.5} & \textbf{25.4} & \textbf{31.8} & \textbf{38.04} \\
\bottomrule
\end{tabular}%
}
\end{table}

\subsection{Main Results}
\label{sec:sota}

Table \ref{tab:main_results} presents comparative results on the GeoText-1652 dataset under both zero-shot and fine-tuned settings. In the zero-shot setting, all methods exhibit limited performance due to the significant domain gap between natural images and drone-view aerial imagery. XVLM achieves the best zero-shot performance among baselines, benefiting from its multi-grained vision-language pretraining.

In the fine-tuned setting, GRASP achieves competitive performance across all metrics. Compared to the GeoText-1652 baseline \cite{chu2024towards}, GRASP improves R@1 by a significant +3.3\,\% for Image Query and +0.9\,\% for Text Query. The larger improvement on Image Query demonstrates that GRASP successfully overcomes the visual isomorphism bottleneck inherent in overhead drone views. While baseline models frequently retrieve visually similar but semantically incorrect buildings (due to identical top-down geometric shapes), the integration of SPEM forces the model to heavily penalize mismatching fine-grained attributes (e.g., roof color, localized materials). This directly addresses the core challenge of aerial imagery. We also compare against NAS \cite{wang2025enhancing}, which explores feature-space negative construction. While effective in general vision tasks, its performance gains are limited in drone-view scenarios (+0.1\,\% mR over baseline). In contrast, GRASP leverages the proposed RFA and SPEM for fine-grained drone-view cross-modal understanding, yielding substantially better results (+1.99\,\% mR over baseline).

\subsection{Ablation Studies}
\label{sec:ablation}

To analyze the contribution of each component, we conduct ablation experiments by enabling different combinations of RFA and SPEM components (Table \ref{tab:ablation_with_mr}). Starting from the baseline (mR 36.05\,\%), applying SPEM with semantic perturbations (TSP+VSP) improves performance to 37.01\,\%, validating that perturbation-driven hard negatives and prototype-based augmentation enhance fine-grained discrimination even without explicit RFA. In parallel, enabling RFA via $\mathcal{L}_{\mathrm{rank}}$ and $\mathcal{L}_{\mathrm{RITC}}$ yields 36.75\,\%, indicating region-aware constraints help suppress background noise and strengthen separability.

When combined, RFA further complements SPEM (e.g., 37.30\,\% / 37.26\,\% with additional ranking or region-level contrastive constraints), and under RFA, VSP provides a larger gain than TSP (37.53\,\% vs. 37.33\,\%), highlighting that explicitly modeling the visual attribute manifold is particularly vital when vertical geometric cues are lost in aerial perspectives. Ultimately, integrating RFA with both TSP and VSP achieves the best result (38.04\,\%), confirming the powerful synergy between suppressing background clutter (RFA) and enforcing micro-level attribute discrimination (SPEM).

\begin{table}[!t]
\caption{Fine-tuned and zero-shot retrieval on ERA.}
\centering
\label{tab:era_comparison}
% 使用 \linewidth 确保表格适应单栏宽度
\resizebox{\linewidth}{!}{%
\begin{tabular}{l ccc ccc c}
\toprule
\multirow{2}{*}{\textbf{Method}} & \multicolumn{3}{c}{\textbf{Image Retrieval (\%)}} & \multicolumn{3}{c}{\textbf{Text Retrieval (\%)}} & \multirow{2}{*}{\textbf{mR (\%)}} \\
\cmidrule(lr){2-4} \cmidrule(lr){5-7}
 & R@1 & R@5 & R@10 & R@1 & R@5 & R@10 & \\
\midrule
\rowcolor{gray!20}
\multicolumn{8}{c}{\textit{Fine-tuned Results on ERA Dataset}} \\
\midrule
VSE++ & 10.13 & 35.20 & 53.91 & 9.79 & 30.40 & 42.90 & 30.39 \\
PVSE K=2  & 11.04 & 35.57 & 51.65 & 11.31 & 32.60 & 46.95 & 31.52 \\
PVSE K=1  & 11.14 & 36.08 & 53.75 & 9.96 & 33.95 & 47.97 & 32.14 \\
CLIP  & 12.73 & 37.33 & 51.52 & 11.31 & 31.92 & 43.91 & 31.45 \\
PCME  & 13.85 & 42.87 & 60.64 & 14.69 & 35.30 & 49.15 & 36.08 \\
AMFMN-soft  & 14.18 & \textbf{46.79} & 62.87 & 14.35 & 38.01 & 52.02 & 38.04 \\
AMFMN-sim  & 13.75 & 43.41 & 59.59 & 14.02 & 34.12 & 51.52 & 36.06\\
AMFMN-fusion  & 11.62 & 42.26 & 60.51 & 15.20 & 36.99 & 50.33 & 36.15 \\
GALR  & 14.03 & 45.15 & 64.54 & 12.38 & 36.59 & 50.90 & 37.27 \\
VCSR  & 13.69 & 46.31 & \textbf{66.37} & 15.65 & 38.28 & 53.49 & 38.96 \\
\midrule
\rowcolor{gray!20} 
\multicolumn{8}{c}{\textit{Zero-shot Results on ERA Dataset (Fine-tuned on GeoText-1652)}} \\
\midrule
XVLM  & 14.19 & 36.15 & 50.34 & 14.05 & 39.46 & 53.99 & 34.70 \\
GeoText-1652 & 17.91 & 39.19 & 54.73 & 17.09 & 42.30 & 56.76 & 38.00 \\
\textbf{GRASP (Ours)} & \textbf{18.61} & 42.87 & 55.23 & \textbf{19.46} & \textbf{44.39} & \textbf{56.25} & \textbf{39.47} \\
\bottomrule
\end{tabular}%
}
\end{table}

\subsection{Zero-Shot Generalization Evaluation}
\label{sec:zeroshot}

To evaluate the generalization capability of GRASP, we conduct zero-shot experiments on the ERA dataset \cite{huang2024visual}. Models are trained exclusively on GeoText-1652 and directly evaluated on ERA without any fine-tuning, simulating real-world deployment scenarios where the model encounters novel geographic regions and scene types.

As shown in Table \ref{tab:era_comparison}, GRASP obtains mR of 39.47\,\%, outperforming the best fine-tuned method VCSR by +0.51\,\%. Compared to the GeoText-1652 baseline under the same zero-shot setting, GRASP improves mR by +1.47\,\%. The substantial improvements in R@1 for both Image Retrieval (+0.70\,\%) and Text Retrieval (+2.37\,\%) demonstrate that GRASP learns transferable representations that generalize across different aerial image domains. This can be attributed to two factors: (1) RFA promotes domain-robust cross-modal grounding; and (2) SPEM improves transferability by learning domain-invariant texture prototypes.

% This strong zero-shot performance can be attributed to two factors: (1) RFA promotes domain-robust cross-modal grounding by aligning global and region-level representations under explicit margin constraints; and (2) SPEM improves transferability by learning domain-invariant texture prototypes in SPC and applying TSP/VSP during training, which exposes the model to diverse attribute variations.

% These gains indicate that GRASP learns representations transferable across aerial domains.

\Attention

\subsection{Visualizing Attention for Semantic Grounding}
As shown in Figure~\ref{Attention}, to qualitatively evaluate the effectiveness of our GRASP framework in addressing the dual challenges of drone-view retrieval, we visualize and compare the gradient-based class activation maps of GRASP against the baseline GeoText-1652 \cite{chu2024towards}.

As observed in row c, the baseline frequently exhibits diffuse and inaccurate attention. At the macro level, it suffers from Cross-Modal Focus Misalignment, where its activation maps are easily dominated by surrounding environmental noise and often spill over into irrelevant background regions. At the micro level, when confronted with Visual Isomorphism, the baseline struggles to ground the specific descriptive attributes. For example, it fails to accurately isolate the specific "red brick building with a domed roof" (col 2) from neighboring buildings of similar size and shape, indicating an over-reliance on global structural matching rather than precise semantic comprehension.

In contrast, GRASP demonstrates highly localized and accurate semantic grounding (row d), and most crucially, GRASP successfully captures fine-grained textural attributes, overcoming the severe visual homogenization inherent in UAV views. As shown in the visualizations, our method precisely localizes micro-level semantic cues such as a "white building" (col 1) and a "red brick building" (cols 2 and 3). This superior performance can be attributed to the following factors:

First, the integration of RFA effectively suppresses background interference. By explicitly introducing region-level contrastive learning and cross-modal ranking, RFA recalibrates the model's focus from coarse global scenes to object-centric target areas. This allows GRASP to tightly concentrate its activation on the actual foreground subjects rather than the expansive, noisy terrain surrounding them. Second, this powerful discriminative capability is the direct result of our SPEM strategy. By actively constructing counterfactual hard negatives via TSP and VSP, SPEM heavily penalizes subtle attribute mismatches. This mechanism forces the network to look beyond identical geometric shapes and strictly ground the exact fine-grained attributes (e.g., roof color, localized materials) specified in the text.

In summary, the visualizations confirm the synergistic combination of RFA and SPEM effectively transitions the model from background-dominated global alignment to precise, attribute-aware semantic grounding, proving GRASP's robustness in drone views.

\caseStudy

\subsection{Retrieval Case Study}
\label{app:case-study}
To demonstrate the efficacy of our proposed framework in real-world scenarios, we conduct a detailed case study on text-image retrieval. The text query requires the model to identify specific architectural features (``red brick'', ``white windows'', ``white trim'') arranged in a specific spatial layout (``tall buildings on either side of the street''). Figure~\ref{caseStudy} visualizes the retrieval rankings of our method compared to strong baselines.

\paragraph{Baseline:} The baseline models fail to retrieve the ground truth immediately, locating it at Rank 5 and Rank 3, respectively. While the incorrect images retrieved by these models share global contextual similarities (e.g., aerial views of urban areas), they fail to capture the fine-grained features corresponding to the query due to the aforementioned visual isomorphism.
    
\paragraph{Ours:} Our method achieves a higher recall, retrieving the correct image at Rank 1. The visualization confirms that our model strictly captures both the localized textural attributes and the spatial relations. This aligns with our claim that integrating SPEM allows the model to filter out geometrically similar but attribute-mismatched candidates, a capability missed by baseline.